%% file: main.tex
\documentclass[a4paper,fleqn]{cas-dc}

\usepackage[numbers]{natbib}

\usepackage{lineno,hyperref}
\usepackage{amsmath,graphicx}
\usepackage{todonotes,comment,pbox,multirow,pifont,graphics,amsmath,textcomp,subcaption,array,float,adjustbox,mdframed}
\def\tsc#1{\csdef{#1}{\textsc{\lowercase{#1}}\xspace}}
\tsc{WGM}
\tsc{QE}
\tsc{EP}
\tsc{PMS}
\tsc{BEC}
\tsc{DE}

\begin{document}
\sloppy
\let\WriteBookmarks\relax
\def\floatpagepagefraction{1}
\def\textpagefraction{.001}
\shorttitle{Investigations on Speech Recognition Systems for Low-Resource Dialectal Arabic-English Code-Switching Speech}
\shortauthors{Hamed et~al.}

\title [mode = title]{Investigations on Speech Recognition Systems for Low-Resource Dialectal Arabic-English Code-Switching Speech}
\author[IMS]{Injy Hamed}
\ead{injy.hamed@ims.uni-stuttgart.de}
\cormark[1]
\author[IMS]{Pavel Denisov}
\ead{pavel.denisov@ims.uni-stuttgart.de}

\author[IMS]{Chia-Yu Li}
\ead{chia-yu.li@ims.uni-stuttgart.de}

\author[RAISA]{Mohamed Elmahdy}
\ead{melmahdy@raisaenergy.com}

\author[GUC]{Slim Abdennadher}
\ead{slim.abdennadher@guc.edu.eg}

\author[IMS]{Ngoc Thang Vu}
\ead{thang.vu@ims.uni-stuttgart.de}

\address[IMS]{Institute for Natural Language Processing, University of Stuttgart, Germany}
\address[RAISA]{Data Science Department, Raisa Energy LLC, Cairo, Egypt}
\address[GUC]{Computer Science Department, The German University in Cairo, Egypt}

\begin{abstract}
  Code-switching (CS), defined as the mixing of languages in conversations, has become a worldwide phenomenon. The prevalence of CS has been recently met with a growing demand and interest to build CS ASR systems. In this paper, we present our work on code-switched Egyptian Arabic-English automatic speech recognition (ASR). We first contribute in filling the huge gap in resources by collecting, analyzing and publishing our spontaneous CS Egyptian Arabic-English speech corpus. We build our ASR systems using DNN-based hybrid and Transformer-based end-to-end models. In this paper, we present a thorough comparison between both approaches under the setting of a low-resource, orthographically unstandardized, and morphologically rich language pair. We show that while both systems give comparable overall recognition results, each system provides complementary sets of strength points. We show that recognition can be improved by combining the outputs of both systems. We propose several effective system combination approaches, where hypotheses of both systems are merged on sentence- and word-levels. Our approaches result in overall WER relative improvement of 4.7\%, over a baseline performance of 32.1\% WER. In the case of intra-sentential CS sentences, we achieve WER relative improvement of 4.8\%. Our best performing system achieves 30.6\% WER on ArzEn test set.
\end{abstract}

\begin{keywords}
speech recognition, speech corpus, code-switching, Arabic-English, low-resource, end-to-end ASR, DNN-based ASR, system combination
\end{keywords}

\maketitle

\setcode{utf8}
\input{sections/intro}

\input{sections/rw}
\input{sections/arzen}
\input{sections/experiments}
\input{sections/results}
\input{sections/qualitative_analysis}
\input{sections/merging_hyp}

\input{sections/conclusion_v2}
\section*{Acknowledgements}
This work was supported by the DAAD (German Academic Exchange Service). We would like to thank Nizar Habash, Özlem Çetinoğlu, and Mohamed Balabel for their valuable suggestions and linguistic advice. We would also like to thank the reviewers for their insightful comments and constructive feedback.
\appendix

\input{sections/intrawordCS_table}
\bibliographystyle{cas-model2-names}

\bibliography{main}

\end{document}

%% file: sections/intro.tex
\section{Introduction}
It is common for speakers in multilingual societies to alternate between languages in conversations, a multilingual phenomenon, referred to as ``code-switching'' (CS). In the middle east, colonization and international businesses and education have played a major role in introducing English and French into conversational language. CS is prevalent in Arab countries such as Arabic-French in Morocco \cite{Ben83} and Algeria \cite{BD83}, Arabic-English in Egypt \cite{Abu91}, Saudi Arabia \cite{OI18}, Jordan \cite{MA94}, Kuwait \cite{Akb07}, Oman \cite{AlQ16} and UAE \cite{Khu03} and a high level of multilingualism is found in Lebanon \cite{BB11} and Tunisia \cite{Bao09} with the mixing of Arabic and both English and French. CS can occur at the levels of sentences, words, and morphemes in the case of morphologically rich languages, such as Arabic. In the following examples, we present the different CS types. As Arabic follows a right-to-left writing system, we use arrows to denote the starting direction of a sentence.
\begin{itemize}
    \item Inter-sentential CS: defined as switching languages from one sentence to another. For example:\\
    ``\textrightarrow It was really nice. \<اتعلمت كتير> .''  \newline(It was really nice. I learnt a lot.)
    \item Extra-sentential CS: where a loan word is borrowed from the secondary language. For example:\\
    ``project \<انا خلصت ال > \textleftarrow'' \newline(I'm done with the project).
    \item Intra-sentential CS (code-mixing): defined as using multiple languages within the same sentence. For example:\\
    ``\textrightarrow I do not think \<أني عاوز أبقى> student anymore.''  \newline(I don't think I want to be a student anymore.)
    \item Intra-word CS (morphological CS): where switching occurs at the level of morphemes. For example:\\
    ``.[conference \<ال>] 
    [target \<هن>]
    \<احنا> \textleftarrow'' 
    \newline(We [will target] [the conference])
\end{itemize}

The global prevalence of this multilingual phenomenon has placed great demand over ASR systems to be able to handle such mixed input. Despite the increasing attention CS ASR has received in the past decade \cite{BK12,VLW+12,AAL17,YBW+18,SS18,SSS+18,YLY+19,LV19,LLY+19,CSE+20}, it is still considered to be in a rather fledgling state, with only few language pairs covered, and a huge space for research investigations and advancements. One of the main bottlenecks hindering the development of CS NLP applications is the lack of resources \cite{CV16}. It is usually the case that CS data is scarce, with resources only covering few language pairs, and are usually very limited in size. In the case of Arabic CS, the diglossic nature of the Arabic language magnifies this problem even more. Besides the Modern Standard Arabic (MSA), which is the formal Arabic that is taught in schools and acts as a linguistic unifying force to all Arabic speakers, each country has its own dialect (and sub-dialects), which is used in daily communication. Since CS is language-dependent, a CS corpus for each dialect is needed covering the foreign languages associated with this dialect. Few corpora are available for Arabic CS, where Egyptian Arabic-English suffers from lack of speech corpora. In order to build our Egyptian-Arabic ASR system, our first task was to collect our ArzEn CS speech corpus. 
The corpus is described in Section \ref{sec:arzen}.\\

Most of the previous work on CS ASR has used hybrid ASR systems \cite{LV19,LLY+19,YLY+19}; following the traditional acoustic modeling, language modeling and pronunciation dictionary (lexicon) pipeline. 
Recently, E2E systems are preferred by researchers for their simplicity and success in monolingual \cite{CSW+18} and multilingual \cite{TSW+18} settings. While E2E systems have been previously investigated for CS ASR \cite{LJZ+18,ZKP+18,SWW+19,LV19,WMC+18,LLY+19,YLY+19}, their competency versus DNN-based hybrid systems is still questionable, where literature still lacks a thorough comparison between both systems for the low-resourced CS ASR task. In this work, we present our work on building Arabic-English CS ASR systems using both E2E and DNN-based hybrid models. In Section \ref{sec:exp_results}, we present a thorough comparison between both systems with regards to the extent of multilingual as well as crosslingual knowledge transfer achieved by both systems, their tolerance to unstandardized orthography, and their reliance on CS data for the success of cross-lingual knowledge transfer from monolingual to CS context and the success of transfer learning in E2E systems.\\

Building ASR systems for CS Egyptian Arabic-English is a rather challenging task. From the dialectal Arabic (DA) side, the system is challenged by two main DA issues: unstandardized orthography and morphological richness. From the English side, despite the fact that there are huge amounts of English speech corpora available, our system can only make use of a few hours of such data. This restriction is placed as a result of our primary language (Egyptian Arabic) being a low-resource language, a limit is then consequently placed on the amount of secondary language (English) data that can be used in the models, after which addition of further data harms recognition. We show that DNN-based hybrid systems show superiority in English recognition. The problem of limited English training data is alleviated with the use of lexicons which act as a valuable knowledge resource, enabling the system to recognize out-of-vocabulary (OOV) and infrequent words. However, the reliance on lexicons limits the performance of DNN-based in the case of DA, where they fall short against the challenges of unstandardized orthography and morphological richness. E2E systems on the other hand, as their performance is not restricted with lexicons, show superiority in DA recognition.\\

Motivated by our findings and the fact that system combination is found to be most effective in the case of ASR systems that provide similar performance and are complementary in their errors \cite{HKS+06}, we apply system combination to utilize the strengths of both systems. 
Combining hypotheses of ASR systems has been previously explored, where researchers investigated several techniques including Recognizer Output Voting Error Reduction (ROVER) \cite{Fis97,LSS+02}, \textit{i}ROVER \cite{HHO+07}, N-best ROVER \cite{SBB+00}, Confusion Network Combination (CNC) \cite{EW00}, \textit{i}CNC \cite{HSN08}, Bayesian model combination (BAYCOM) \cite{San05}, lattice rescoring \cite{MKS+10}, and lattice combination \cite{XPM+10,LSS02}. 
An overview on previous work for ASR system combination is presented in \cite{Bre08}. 
Up to our knowledge this work represents the first efforts towards combining outputs of hybrid and E2E systems for improving CS ASR. 
In our work, we present three approaches for combining outputs from both systems on the sentence- and word-levels. We show how such approaches can be used to leverage the strengths of both systems to reach a better list of hypotheses and significantly improve CS ASR.\\
    
In the scope of this work, we make the following contributions:
\begin{itemize}
    \item Release our ArzEn corpus to motivate further research in this direction. The corpus is primarily designed for ASR, however, it also serves as a useful resource for other NLP tasks as well as linguistic, sociolinguistic and psycholinguistic CS studies.
    \item Present the first ASR system for Egyptian Arabic-English speech, while reflecting on language-specific challenges and CS ASR limitations. Our final system achieves a WER of 30.6\% and CER of 18.7\% on test set.
    \item Present a thorough comparison between E2E and DNN-based hybrid systems.  
    While the performance of E2E systems is usually tied to the availability of huge amounts of training data, we show that they provide comparable results to DNN-based hybrid systems in our low-resource setting. 
    Through our error analysis, we show that each system brings its own set of strengths. The superiority of E2E systems in Arabic recognition is attributed to their tolerance to DA issues, which greatly affect the performance of hybrid systems due to their reliance on lexicons. In the case of English recognition, systems suffered from the restricted amount of English training data. While this problem greatly affected E2E systems, DNN-based hybrid systems performed significantly better as knowledge was gained from lexicons. The performance of the E2E system is further improved using transfer learning.
    \item Propose effective approaches for combining outputs of E2E and hybrid systems.  The best performance is achieved by a hybrid system combination, where hypotheses are merged on both sentence- and word-levels. 
    Our approach leverages the flexibility of E2E systems with regards to DA recognition as well as the strengths provided by the DNN-based systems lexicon supporting English recognition and infrequent and OOV words.
    Our results confirm systems' complementarity, 
   where system combination results in an overall WER relative improvement of 4.8\% in the case of intra-sentential CS.

\end{itemize}

%% file: sections/rw.tex
\section{Related Work}
In this section we give an overview on how CS ASR systems have evolved over the years, we present the current work done on CS ASR for the Arabic language, as well as currently available CS speech corpora.

\subsection{CS ASR}
Over the past years, CS ASR systems have remarkably evolved. The first efforts included a multi-pass approach \cite{LL08}, where a language boundary detection (LBD) algorithm is used to divide the input utterance into segments that are language-homogeneous. The language identity of each segment is identified using a language identity detection (LID) algorithm. The corresponding language-dependent ASR system is then used. While this approach is suitable for recognizing intersentential CS, it does not perform well for intrasentential CS. Researchers then opted for a one-pass approach, where one multilingual ASR system -consisting of multilingual acoustic model, pronunciation dictionary and language model- is capable of recognizing such mixed speech. The majority of the work done was based on the GMM-HMM model \cite{BK12,VLW+12}. With the current advances in deep learning, researchers have directed their work towards DNN-based approaches, which have proven to outperform the traditional GMM-HMM models \cite{SS18,SSS+18,YBW+18}. While hybrid systems have achieved great success, they require language-specific lexicons (which may not be present for low-resource languages) and linguistic knowledge. E2E approach, on the other hand, has simplified the training process by removing the hand-engineered components requiring human expertise. E2E ASR systems have proved to be as successful as the hybrid ASR systems for resource-rich languages \cite{HWZ+17}. However, the limit of acoustic resources hinders the performance of E2E CS ASR systems more severely than hybrid ASR systems \cite{YLY+19}. While a number of researchers have investigated the use of E2E systems for CS ASR, literature still lacks a thorough comparison between E2E and DNN-based approaches under the low-resource CS setting. Moreover, only few language pairs have been investigated so far, including Mandarin-English \cite{LV19,LJZ+18,ZKP+18,SWW+19}, Chinese-English \cite{LLY+19,WMC+18}, and Frisian-Dutch \cite{YLY+19}. Research lacks such an investigation for CS Egyptian Arabic-English language pair.

\subsection{Arabic CS ASR}
Despite Arabic being one of the most popular languages (ranking $6^{th}$ according to total number of speakers \cite{ESF19}) and the prevalence of CS across Arab countries \cite{AML16,HEA18,AAL18}, there is a huge research gap in the field of the Arabic CS ASR. While many researchers have worked on ASR for Modern Standard Arabic \cite{EEA+03,KSS+11,MKC+11,AZC+14,CAD+14,ABG+16} and dialectal Arabic \cite{KBD+03,KVB+06,SMB11,EGM12,ABG+16,AVR17,NHA+17,Ali18,AR18,ASS+19}, work on CS Arabic ASR is still in its initial stages. In \cite{BMC08}, Bayeh et al. presented an ASR system for Arabic Broadcast News (BN), where they rely on crosslingual techniques to allow the baseline systems trained on MSA data to recognize dialectal languages (mainly Levantine/Maghrebian) as well as French embedded language. Barik and Lestari \cite{BL19} developed an ASR system for Indonesian-Arabic, where Arabic is embedded into Indonesian conversations. The system's WER was improved by including lexicon entries for Arabic words, adding CS sentences in the text corpus used for language modeling (LM) training, and merging phonemes using International Phonetic Association (IPA). Elfahal \cite{Elf19} worked on developing CS ASR system for read Sudanese Arabic-English speech. The authors built the ASR system using mixed LM, mixed phonetic dictionary and GMM-HMM acoustic model. The output of the ASR system, along with the hypotheses confidence scores were then passed to a second stage where the language of each word is identified using lexicon look-up. In \cite{CSE+20}, Chowdhury et al. present their work on dialectal Arabic-MSA ASR system. Also related to the scope of our research is the work of Amazouz et al. \cite{AAL17}, where the authors do not work on ASR, but rather on language identification in speech for the Algerian Arabic-English language pair. Our system presents the first work for the Egyptian Arabic-English language pair.\\

\subsection{CS Speech Corpora}
The vast majority of the available CS speech corpora have covered Chinese-English \cite{CCC+09,SWY+11,LYF12,LTC+15,AT12}, Hindi-English \cite{DF14,RS17,SSS+18,SDS18,PSG17} and Spanish-English \cite{SL08,DDH+14,RS17} language pairs. Less work has covered Arabic-English \cite{Ism15,HEA18,HVA20}, Arabic-French \cite{AAL18,MML16}, Frisian-Dutch \cite{YAK+16}, Mandarin-Taiwanese \cite{LLC+06,LL08}, Turkish-German \cite{Cet17}, English-Malay \cite{AT12}, English-isiZulu \cite{WN16} and Sepedi-English \cite{MDD13}. For Arabic CS, speech corpora are very limited in size as well as covered language-pairs. Ismail \cite{Ism15} collected an 89-minutes speech corpus by recording and transcribing informal dinner gatherings involving 6 participants. Amazouz et al. \cite{AAL18} gathered a 7.5 hours of read speech from books and movie transcripts, as well as informal conversations, gathered from 20 speakers. The corpus contains transcriptions, sentence segmentation, language boundary and phone-level time codes information. A  Maghrebian Arabic-French speech corpus \cite{MML16} was also collected by Amazouz et al. containing 53 hours of spontaneous speech gathered from TV entertainment and talk shows, involving speakers from Algeria, Morocco and Tunisia. The corpus contains sentence segmentation and language annotation. For Egyptian Arabic, several corpora were collected in the scope of our project. In \cite{HEA18}, we present a 6-hours corpus of informal interviews with 12 speakers in technical domain, where transcriptions were obtained for 4 hours. The corpus is still under processing to be used in ASR, where the remaining transcriptions are being obtained and the whole corpus is being segmented. In \cite{HVA20}, we present a 12-hours speech corpus that is transcribed, sentence-segmented, and annotated for speaker information. Information about the corpus is given in Section \ref{sec:arzen}.

%% file: sections/arzen.tex
\section{ArzEn Speech Corpus}
\label{sec:arzen}

\begin{figure*}[pos=h]
     \centering
     \begin{subfigure}[b]{\textwidth}
         \centering
         \includegraphics[width=\linewidth]{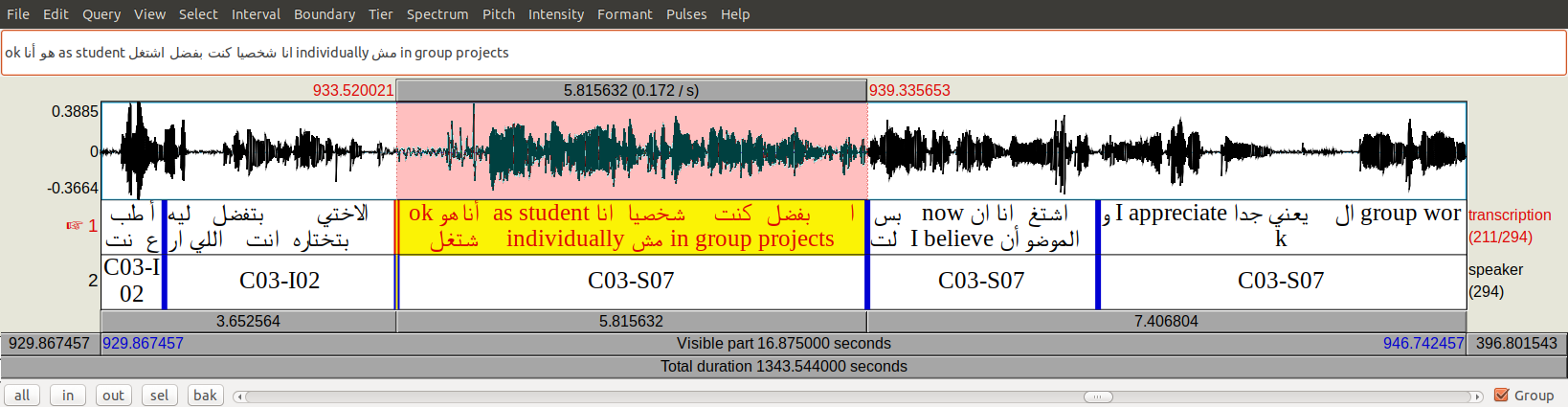}
         \caption{Using Praat for the annotation process.}
         \label{fig:praat}
     \end{subfigure}
     \hfill
     \begin{subfigure}[b]{0.7\textwidth}
         \centering
         \begin{mdframed}
         \includegraphics[width=\linewidth]{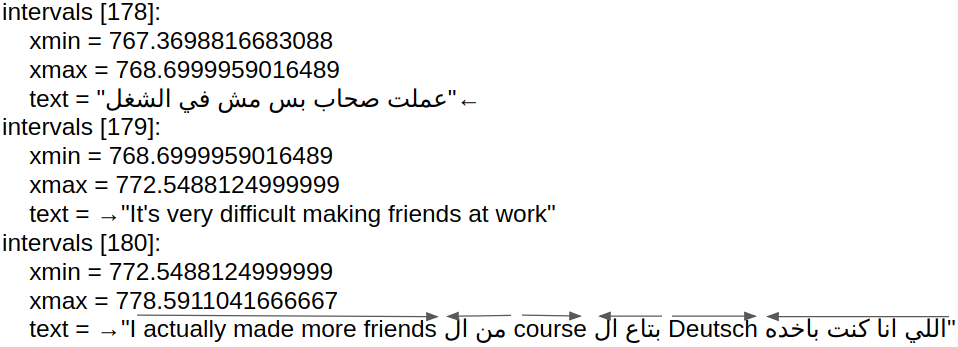}
         \end{mdframed}
         \caption{Example of the TextGrid output from Praat for the transcription tier. The arrows beside the sentences show the sentence starting direction, while the arrows on top of the words show the direction of the language-homogeneous blocks.}
         \label{fig:corpus_sample}
     \end{subfigure}
        \caption{Corpus transcription and annotation process.}
        \label{fig:corpus_transcription_annotation}
\end{figure*}

In this Section, we will briefly present the ArzEn\footnote{Arz and En are the codes for Egyptian Arabic and English in Ethnologue.} speech corpus, discussing the collection and annotation process as well as corpus code-switching statistics. We extend our work in \cite{HVA20} with: (1) releasing the corpus \footnote{\url{https://github.com/DigitalPhonetics}} and (2) improved annotation and further analysis with regards to intra-word CS in conversational Egyptian Arabic speech.\\

ArzEn is a 12-hours Egyptian Arabic-English CS spontaneous speech corpus. The corpus is collected through informal interviews with 38 Egyptian bilingual speakers, who are fluent in English. The interviews were held at The German University in Cairo (GUC), a private university where English is the instruction language. Participants are in the age range of 18-35, where 61.5\% are males and 38.5\% are females. All participants are students (55\%) or employees (45\%) of the GUC. The interviewees were asked questions covering several topics, including education, personality, personal life, career, hobbies, traveling, work and life experiences, role model and technology. To ensure good audio quality, all recordings were carried out in a soundproof room and at a sampling rate of 48kHz. 
The corpus is mainly designed to be used in Automatic Speech Recognition (ASR) systems, however, the interview setup is designed so that the corpus acts as a useful resource for analyzing the CS phenomenon from linguistic, sociological, and psychological perspectives. Interviews included one interviewee and two interviewers; a male and a female. 
The set of questions as well as the way the questions were asked were also fixed throughout the interviews. 
After the interview, the participant was asked to fill in three forms:
\begin{itemize}
    \item Questionnaire: gathering information about the participants including their demographics, education, work, travelling experiences, and their perceptions and opinions regarding code-switching. 
    \item Big Five Personality Test \cite{Gol92}: assesses five major dimensions of personality: Openness, Conscientiousness, Agreeableness, Extraversion, and Neuroticism. The Big Five Personality test was chosen as it is the most widely used and extensively researched model of personality \cite{GRS03} and because it consists of 50 questions, thus only requires around 10 minutes from the participants.
    \item Self-Assessment Manikin Test \cite{BF17}: a picture-oriented questionnaire developed to measure emotional response.
\end{itemize}

These forms were used in order to gather more information about the interviewees, motivating further socio- and psycho-linguistic research to be conducted for this under-resourced CS language pair.
\subsection{Corpus Transcription and Annotation}

The corpus was transcribed by professional transcribers, and revised by one of the authors, who is a fluent Arabic-English bilingual. In order to address the unstandardized issue of Dialectal Arabic orthography, we based our transcription guidelines on the conventions developed and used by the Egyptian Arabic Wikipedia community  \footnote{\href{https://arz.wikipedia.org/wiki/\%D9\%88\%D9\%8A\%D9\%83\%D9\%8A\%D8\%A8\%D9\%8A\%D8\%AF\%D9\%8A\%D8\%A7:Introduction_in_English\#Rules_of_writing}{https://arz.wikipedia.org/wiki/\<ويكيبيديا>:Introduction\_in\_English\newline\#Rules\_of\_writing}}. In the case of ambiguous cases where several orthographic varieties are allowed, we made decisions to restrict the number of possibilities to usually one variant only. We use the following tags for non-speech parts: [HES] for hesitation, [HUM] for humming, [COUGH], [LAUGHTER], and [NOISE]. Sentence segmentation was performed such that each segment spans of maximum 25 seconds, and each segment was annotated with the speaker ID. 
Praat\footnote{\url{http://www.fon.hum.uva.nl/praat/}} is used for annotation, where we use two tiers: transcription and speaker. Figure \ref{fig:praat} shows an example of the annotation process. Figure \ref{fig:corpus_sample} shows the TextGrid file produced by Praat. Since Arabic adopts a right-to-left writing system, reading code-mixed Arabic-English sentences could be confusing as language-homogeneous segments are read in different directions. Also, the starting direction of the sentence, depends on the language of the first word, where sentences beginning with an Arabic character start from the right side, while sentences beginning with a Latin character start from the left side. In order to make it easier throughout the paper, we will use a small arrow to denote the starting direction of sentences, and we will place arrows on top of words to guide the reader through the word sequences. For example, in Figure \ref{fig:corpus_transcription_annotation}, the text in the third interval starts from the left side and is read in the following order: 
I - actually - made - more - friends - 
\<من>
-
\<ال>
-
course 
-
\<بتاع>
-
\<ال>
-
Deutsch
-
\<اللي>
-
\<انا>
-
\<كنت>
-
\<باخده>.

\subsection{Corpus Statistics}
In Table \ref{table:arzen_stats_overall}, we provide an overview on the corpus, presenting the number of speakers, speech and non-speech duration, average sentence length and duration, and average speaking rate, as well as the number of sentences, words, and tokens. The difference between the number of words and tokens is due to intra-word CS, where a compound word consisting of Arabic affixes and an English word is treated as separate tokens. In Table \ref{table:arzen_stats_cs}, we report the level of code-switching available in the ArzEn corpus. In order to provide a thorough CS analysis on the ArzEn corpus, we report code-switching metrics reflecting all 4 types of code-switching.
\begin{table}[pos=h]
    \begin{subtable}{\linewidth}
        \centering
    \begin{tabular}{|p{6cm}|p{1.5cm}|}
      \hline
      \textbf{Category}&\textbf{Value}\\
      \hline
      \# Speakers & 40\\
      \hline
      \# Interviewers & 38\\
      \hline
      \# Interviewees & 2\\
      \hline
      Average Interview Duration & 0.32 hours\\
      \hline
      Total Duration &12 hours\\
      \hline
      Speech Duration &11.4 hours\\
      \hline
      Non-speech Duration &0.7 hours\\
      \hline
      \# Sentences & 6,290\\\hline
      \# Words & 99,340\\\hline
      \# Tokens&102,332\\\hline
      Sentence Duration Range (s)&0.3-24.9\\\hline
      Av. Sentence Duration (s)&6.6\\\hline
      Sentence Length Range (words)&1-95\\\hline
      Av. Sentence Length (words)&16.3\\\hline
      Average speaking rate (words per minute)&149.1\\\hline
    \end{tabular}
       \caption{Overall statistics}
       \label{table:arzen_stats_overall}
    \end{subtable}
    \hfill
    \begin{subtable}{\linewidth}
        \centering
        \begin{tabular}{|p{6cm}|p{1.5cm}|}
      \hline
      \textbf{Category}&\textbf{Value}\\
      \hline
      \multicolumn{2}{|c|}{\textbf{Inter-sentential CS}} \\\hline
      \% Monolingual Arabic Sentences&33.1\%\\\hline
      \% Monolingual English Sentences&3.7\%\\\hline
      \multicolumn{2}{|c|}{\textbf{Intra-sentential CS}} \\\hline
      \% CS Sentences&63.2\%\\\hline
      \% Arabic Tokens&84.9\%\\\hline
      \% English Tokens&15.1\%\\\hline
      \multicolumn{2}{|c|}{\textbf{Morphological CS}} \\\hline
      \% Sentences with morphological CS&28.4\%\\\hline
      \% CS sentences with morphological CS&45.0\%\\\hline
      \% Arabic Words & 84.9\%\\\hline
      \% English Words & 12.3\%\\\hline
      \% Morphological CS words & 2.7\%\\\hline
      \# Morphological CS Words & 2,684\\\hline
    \end{tabular}
        \caption{Code-switching statistics}
        \label{table:arzen_stats_cs}
     \end{subtable}
     \caption{Corpus statistics.}
     \label{table:stats}
\end{table}

\subsubsection{Inter-sentential CS:}
Inter-sentential CS is seen in the corpus, however, with very low frequency, where pure monolingual English sentences only constitute 3.7\% of the sentences. However, it is seen that some sentences are mainly in English but contain few Arabic conjugations. For example: ``I would say probably martial arts \<أو> (or) tennis.''

\subsubsection{Intra-sentential and Extra-sentential CS:}
A total of 63.2\% of the sentences are code-mixed, where 88.9\% are mainly in Arabic with English embeddings, while 7.4\% have more English than Arabic words. On average, in each CS sentence, there are 1.98 switches from Arabic to English and 1.91 switches from English to Arabic. Among the CS sentences, 80.5\% of the sentences start in Arabic and 19.5\% start in English. Throughout the corpus, there are 15,481 English words, which are 15.1\% of the total words. Among the CS sentences, 18.8\% of the words are in English. We also report the Code-Mixing Index (CMI) \cite{DG14}, which represents the level of mixing between languages in text. CMI is defined as:
\[CMI=\frac{\sum_{i=1}^{N}(w_i)-max\{w_i\}}{n-u} \]
where $\sum_{i=1}^{N}(w_i)$ is the total number of words over all languages, $max(w_i)$ is the highest number of words across the languages, $n$ is the total number of words, and $u$ is the total number of language-independent words. Monolingual sentences would have a CMI of 0 and sentences with equal word distributions across languages would have a CMI of $n/N$, which is 0.5 in the case of bilingual utterances. 
The CMI is 0.12 over the whole corpus and 0.17 over CS sentences.

\begin{table*}[pos=h]
\centering
\setlength\tabcolsep{1.5pt}
\begin{adjustbox}{max width=\textwidth}
\begin{tabular}{|c|c|c|c|c|c|r|r|r|r|}
\hline
\textbf{Set}& \textbf{Interviews}& \textbf{F}& \textbf{M}&\textbf{Size(h)} &\textbf{wpm}&\textbf{CMI}&\textbf{\%CS Snt.}&\textbf{\%En Snt.}&\textbf{\% En words}\\\hline
\textbf{Train}&12&4+1&8+1&5.6&156.9&0.14&68.8\%&3.9\%&16.0\%\\\hline
\textbf{Dev}&13&5&8&2.9&143.9&0.14&70.0\%&2.6\%&14.8\%\\\hline
\textbf{Test}&13&5&8&2.9&148.7&0.14&68.7\%&2.1\%&17.0\%\\\hline
\end{tabular}
\end{adjustbox}
\caption{Overview on training, development and test sets. For each set, we show the number of interviews, gender distribution, duration, word per minute rate (wpm), CMI as well as the percentages of CS sentences, English sentences, and English words.}
\label{table:sets}
\end{table*}

\subsubsection{Intra-word CS:}
We extend our analysis presented in \cite{HVA20} by improving our annotation for morphological CS. In \cite{HVA20}, no annotation was done for intra-word CS, where English words and Arabic affixes/clitics were separated with spaces, for example ``\<ال> TASK \<ات>'' (the tasks). In this paper, We improve our annotation for morphological CS words as follows: \textbf{Arabic prefixes/proclitics + English words \# Arabic suffixes/encl-itics}, for example ``\<ال>+TASK\#\<ات>''. Such an annotation allows further investigations on intra-word CS in the case of Egyptian Arabic-English language pair, and allows us to measure the recognition improvements in our ASR experiments specifically on this type of CS. 
We extend our corpus with morphological CS analysis. The corpus contains 2,684 morphological CS words. In Table \ref{table:arzen_affixes}, we present the list of Arabic clitics and affixes in the context of morphological CS, as well as their frequencies. Morphological CS occurs most frequently in the form of Arabic prefixes/proclitics attached to English words, where this form is seen 2,440 times. The attachment of Arabic suffixes/enclitics to English words is less frequent, occurring only 14 times. The attachment of Arabic circumfixes to English words is present 10 times. Other combinations of Arabic prefixes/proclicits and suffix-es/enclitics attached to English words occur 7 times, such as
\<ال>+ROBOT\#\<ات> (the robots),
\<لل>+TASK\#\<ات> (to the tasks),
\<ا>+ENHANCE\#\<ه> ((I) enhance it),
\<ى>+PROPAGATE\#\<ها> ((he) propagates it),
\<بى>+EVALUATE\#\<نى> ((he) evaluates me).

\subsection{Adaptation, Development, and Test Sets}
The corpus is divided into train, dev and test sets. We take into consideration having balanced dev and test sets in terms of gender distribution, number of interviews, duration, word per minute (wpm) and CS metrics where we consider the CMI, the percentage of code-mixed sentences, and the percentage of English words. In Table \ref{table:sets}, we present the figures across training, development, and test sets. In order to avoid including the interviewers as common speakers across all sets, their utterances are placed in the train set. Therefore, the train set includes 4 female and 8 male interviewees, in addition to 1 female and 1 male interviewer.

%% file: sections/experiments.tex
\section{Experimental setup}
\input{sections/data}

\subsection{Hybrid ASR}
The hybrid ASR systems were built using Kaldi toolkit \cite{PGB+11}. In this Section, we discuss the multilingual acoustic model, pronunciation dictionary and language model.

\subsubsection{Acoustic Modeling} 
We train hybrid systems using both GMM-HMM and TDNN-HMM \cite{PPK15} architectures. TDNN-DNN was chosen as it has shown superiority over other DNN-based architectures \cite{YBW+18}. The GMM-HMM model is used as point of reference. For the \textbf{GMM-HMM} acoustic model,  the Egyptian Callhome Kaldi recipe was used, which follows a Triphone+SGMM+SAT+fMLLR pipeline. The recipe uses 5 triphone training with 30k Gaussians using 13 Mel-frequency cepstral coefficients (MFCC). The acoustic features are extracted by applying Hamming windowing with a frame length of 25 ms and frame shift of 10 ms. For the \textbf{TDNN-HMM} acoustic model, we use the "s5c" TDNN Switchboard recipe "tdnn\_7q", using the same configurations for training parameters. The alignments produced by the GMM-HMM are used for TDNN training. The default MFCC features combined with i-vectors for speaker adaptation \cite{Sao18} are used, in addition to Standard 3-way speed perturbation \cite{KPP+15} with 0.9, 1.0 and 1.1 speed ratios for 3-fold augmentation of the training data. The TDNN consisted of 15 context splicing layers, each with 1536 nodes, with the following splicing indices:  \{-1, 0, 1\} for layers 2-4, and \{-3, 0, 3\} for layers 6-15. The final softmax output layer computes posteriors for 5,728 triphone states. Motivated by the work of Biswas et al. \cite{BYW+19}, we add 6 CNN layers prior to the TDNN layer, providing a CNN-TDNN architecture. 

\subsubsection{Pronunciation Dictionary}
The Egyptian Colloquial Arabic lexicon \cite{KGA+97}, containing 42 phones and 56,913 words was used for Egyptian Arabic. Librispeech lexicon \cite{PCP+15}, containing 70 phones and 206,508 words was used for English. It is to be noted that Librispeech lexicon is not the most appropriate choice as it does not take into consideration accented Egyptian English, however it was used due to the absence of an Egyptian English lexicon. Phone merging was performed based on linguistic knowledge of a bilingual speaker, where 26 similar phones were merged. This method has shown to outperform multiple data-driven methods \cite{SSS+18}. The lexicon was filtered to words that appear in the speech and text corpora. The OOV rates are 11.0\% (dev) and 11.4\% (test).

\subsubsection{Language Modeling}
For the GMM-HMM models, we use n-gram language modeling using SRILM \cite{Sto02} toolkit. The LM was obtained using interpolation of two models: (1) 4-gram model trained on ArzEn train set transcriptions + extra transcriptions and (2) 3-gram model trained on Callhome and MGB-3 transcriptions, in addition to social media text. Both models used Kneser-Ney smoothing technique. Several interpolation weights were tried and a weight of 0.7 for the first model was found to achieve the best perplexity on the development set. The resulting LM has a perplexity of 566.0 and an OOV rate of 1.1\% on dev set. For CNN-TDNN models, we use recurrent neural network language model (RNNLM). The RNNLM is trained on the same data sets used for training the n-gram LM. The RNNLM consists of 1 LSTM layer with 1000 hidden units. The network is trained for 20 epochs. The perplexity is 208.1 on dev set.

\subsection{E2E ASR}
For E2E systems, there are three popular types: Connectionist Temporal Classification (CTC), Attention Encoder-Decoder (AED), and RNN Transducer (RNN-T). We use joint CTC/attention based E2E framework as it has demonstrated superior performance over CTC and AED separate methods \cite{WHK+17}. The E2E system was trained using ESPnet \cite{WHK+18}. The encoder and decoder consist of 12 and 6 Transformer blocks (layers) with 4 heads, feed-forward inner dimension 2048 and attention dimension 256 \cite{KSWDON+19}. The CTC/attention weight $(\lambda_1)$ is set to 0.3. We applied SpecAugment \cite{PCZ+19} for data augmentation. For decoding, the beam size is 20 and the CTC weight is 0.2. Similar to the CNN-TDNN models, an RNNLM is trained and used in decoding. The RNNLM consists of 1 LSTM layer with 1000 hidden units and is trained for 20 epochs. The perplexity is 131.2 on dev set. Despite our efforts in unifying hyperparameters across systems, there is a difference in perplexities due to implementation differences.

\subsubsection{Transfer Learning}
In our experiments, we first explore multi-condition training, which involves pooling training samples from different conditions and training them simultaneously \cite{DV19}. In multi-condition training, we train one model using pooled CS and monolingual data. We then explore transfer learning, where we train the first model using monolingual data, and pass the parameters to a second model trained on CS data. We also combine both approaches, training the first model using CS and monolingual data, and transferring the parameters to the second model which is solely trained on CS data. In Table~\ref{table:WER}, we report results for multi-condition training, transfer learning, and their combination.

%% file: sections/data.tex
\subsection{Data}
\subsubsection{Speech Corpora}
For CS data, we use ArzEn \cite{HVA20} speech corpus for training, development and testing. For monolingual Egyptian Arabic data, we use CALLHOME \cite{GKA+97} (19.4h-Telephone conversations) and MGB-3 \cite{AVR17} (16h-Egyptian YouTube vid-eos). For the MGB-3 corpus, we excluded utterances having overlapping speech, thus ended up with a subset of 9-hours. For monolingual English data, we use Librispeech \cite{PCP+15} (1,000h-Read audiobooks). For monolingual MSA data, we use MGB-2 \cite{ABG+16} (1,137h-Aljazeera Arabic TV programs (News)). For MGB-2 and Librispeech, we used subsets of the corpora, as outlined later in Section~\ref{sec:exp_results}.

\subsubsection{Text Corpora}
For language modeling, we use the transcriptions of ArzEn train set ($3,346$ sentences, $55,432$ tokens), Callhome ($29,829$ sentences, $213,260$ tokens), and MGB-3 ($3,346$ sentences, $87,093$ tokens), in addition to further CS transcriptions \cite{HEA18}. In order to obtain more text data for CS and monolingual Egyptian Arabic, we leverage data collected from social media platforms ($1,574,208$ sentences, $22,025,671$ tokens) \cite{HZE+19}.

\subsubsection{Data Processing}
Unlike MSA, Dialectal Arabic does not have a standardized orthography, where words can have multiple correct spellings and some letters can be used interchangeably. Regarding the former problem, we identified major differences in the orthography guidelines used in Callhome, ArzEn and MGB-3, and standardized the text in regards to these issues.
Regarding the latter problem, it is common that variants of \textit{Hamzated Alif} (\<أ>, \<إ>, \<آ>) are often written without their \textit{Hamza} and the \textit{Alif-Maqsura} (or \textit{dotless Ya}) (\<ى>) and \textit{dotted Ya}  (\<ي>) are used interchangeably when occurring at the end of a word. This leads to various forms of a word being interpreted as different words, thus higher sparsity and out-of-vocabulary (OOV) rates. In hybrid ASR systems, these inconsistencies affect a word being mapped to its corresponding entry in the lexicon. Therefore, it is important to handle such discrepancies through normalization. In this step, we normalize the textual data and pronunciation dictionary by mapping \textit{Hamzated Alifs} to bare \textit{Alif} (\<ا>) and \textit{dotted Ya} at word end to \textit{dotless Ya}. For the English language, normalization included mapping all letters to uppercase. Punctuation symbols were also removed. 

%% file: sections/results.tex
\section{Experimental Results and Analysis}
\label{sec:exp_results}

\definecolor{Grey}{gray}{0.9}
We compare the two systems (hybrid vs E2E) with regards to: (1) the extent of multilingual and crosslingual knowledge transfer achieved by both systems
(2) their tolerance to orthography un-standardization issue; and (3) their dependence on in-domain CS data for the success of cross-lingual knowledge transfer. 

\subsection{Extent of Multilingual and Crosslingual Knowledge Transfer}
\begin{figure}[pos=h]
     \centering
     \includegraphics[width=\linewidth]{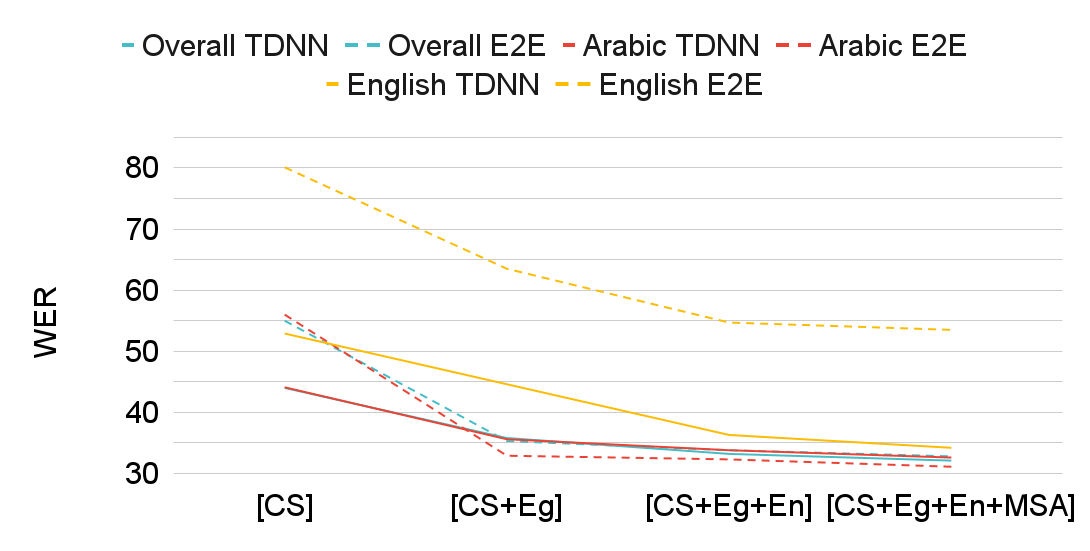}
     \caption{WER achieved by hybrid and E2E systems.}
     \label{fig:WER_multilingual_crosslingual}
\end{figure}

\begin{figure}[pos=h]
    \centering
     \includegraphics[width=\linewidth]{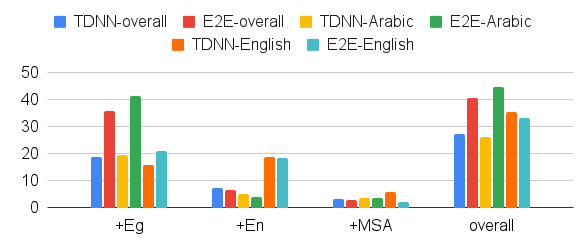}
     \caption{Relative WER improvements achieved by adding Egyptian Arabic (Eg), English (En) and MSA monolingual data.}
     \label{fig:relative_imp}
\end{figure}

In this section, we evaluate hybrid and E2E systems with regards to their ability to transfer knowledge from additional monolingual data, including primary (dialectal Egyptian Arabic), secondary (English), as well as a closely-related (MSA) languages. We examine the effect of adding data from primary, secondary as well as closely-related languages. The corresponding overall as well as language-specific WER figures are presented in Table  \ref{table:WER_knowledge_transfer}. We also present the WER figures as well as relative improvements in Figures \ref{fig:WER_multilingual_crosslingual} and \ref{fig:relative_imp}.\\
\input{sections/results_WER_table_knowledge_transfer.tex}

In Exp2-4, we explore the effect of adding monolingual data for primary and secondary languages. 
CNN-TDNN and E2E models benefit from the addition of monolingual data, while GMM-HMM models show a deterioration in performance. CNN-TDNN models show superiority over E2E systems with limited (5.6 hours) CS data (Exp1). However, E2E systems show higher relative improvements with the addition of monolingual data, and are able to achieve comparable results after adding only 28.4h of monolingual Egyptian Arabic data. As shown in Figure~\ref{fig:relative_imp}, CNN-TDNN and E2E models achieve overall relative improvements of 18.6\% and 35.8\% on test set with the addition of Egyptian Arabic data, and 7.2\% and 6.5\% with English data. As shown in Figure \ref{fig:WER_multilingual_crosslingual}, while the performance of E2E systems is usually tied to the availability of vast amount of training data, they were able to achieve comparable results to CNN-TDNN hybrid systems with only 34h of training data (5.6h CS + 28.4h Egyptian Arabic).\\


In Exp5-11, we investigate the impact of leveraging MSA speech data. While MSA and dialectal Arabic are considered in literature to be separate languages, they share common alphabet and words, presenting a promising source for further acoustic knowledge. Results show that both CNN-TDNN and E2E systems are able to learn from closely-related languages, where the addition of MSA data -up to a certain limit- results in relative improvements of 3.3\% and 3.0\% on test sets, respectively. While the improvements in Arabic recognition can be foreseen (3.6\% and 3.7\% for CNN-TDNN and E2E systems), English recognition also, unexpectedly, improved (5.8\% and 2.2\% for CNN-TDNN and E2E systems). It is to be noted though that the addition of MSA data deteriorates English recognition in some utterances (as shown in Table \ref{table:examples1}-Example 2), however, it provides a general improvement in English recognition for both systems.

\subsection{Tolerance to Un-standardized Orthography}
\input{sections/results_WER_table_normalization}
In order to further compare the performance of both systems, we investigate their tolerance to un-standardization of dialectal Arabic. In Table \ref{table:WER_normalization}, we report the WER achieved by both systems when training models on CS and Egyptian data with (Exp2) and without (Exp12) the pre-processing step of normalization and standardization. While E2E systems show high tolerance to un-standardization where the performance is only slightly affected (0.9\% absolute), hybrid systems suffer from a great (12.0\% absolute) performance deterioration. This can be traced back to their reliance on lexicons, where the lexicon OOV rates improved by 31\% (dev) and 28\% (test) after normalization and standardization. This result reflects:  (1) the reliance of hybrid systems on linguistic knowledge (in addition to linguistic resources) and (2) the flexibility of E2E systems, where corpora following different orthography guidelines can be directly used, without the need for standardization.

\subsection{Reliance on in-domain CS speech data}
\label{sec:in-domain_CS_data}
\input{sections/results_WER_table_CS_data}
\begin{figure}[h]
  \centering
  \includegraphics[width=\linewidth]{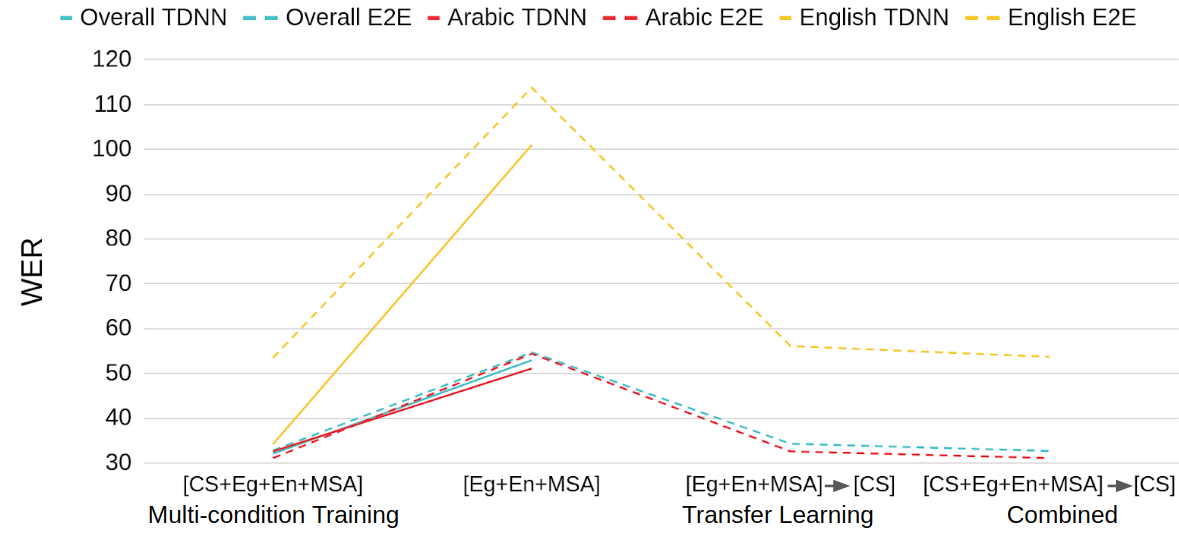}
  \caption{WER of hybrid and E2E systems trained by adding Egyptian Arabic (Eg), English (En) and MSA monolingual data.}
  \label{fig:TL}
\end{figure}

Given the sparsity of CS data and the lack of CS data for many language pairs, We investigate the importance of in-domain CS data and whether we can dispense such time-consuming and expensive data and rely only on monolingual speech data. Most of the work on CS ASR has involved training the ASR systems on CS data. Thus, it is still unclear whether CS ASR systems can be built without CS data, and to what extent is CS data essential for the success of cross-lingual knowledge transfer from monolingual to CS context. We explore the ability of the systems to recognize CS speech, relying only on monolingual data. In Exp13, we train our models using only monolingual data. Although in-domain data constitutes only 12.7\% of the training data, we report that excluding it leads to great performance reductions; 29.6\% (CNN-TDNN) and 32.6\% (E2E) on test set, as shown in Table \ref{table:WER}. While both CNN-TDNN and E2E systems were still able to recognize Arabic speech with 51.1\% and 54.4\% WER  (on test set), they fail to recognize English words, where English WER drops to 101.0\% and 113.8\% (on test set). This demonstrates the necessity of in-domain CS data for the success of cross-lingual knowledge transfer, enabling systems to predict CS switch points. This experiment also highlights the importance of reporting language-specific WER in CS NLP systems.\\

We further extend our experiments to investigate the importance of in-domain CS data in the success of transfer learning in E2E systems. We train an initial model using solely monolingual data, and then fine-tune this pretrained model on CS data (Exp14). Results show that multi-condition training outperforms transfer learning in our task. The problem with the transfer learning approach can be attributed to the low performance of the pretrained model (Exp13-trained only on monolingual data) and its failure with regards to English recognition, as previously discussed, thus limiting the transferred knowledge. Accordingly, we apply transfer learning after multicondition training, as outlined in Exp15, achieving further minor improvements. The effect of incorporating the limited amount of CS data in both approaches on Arabic, English and overall recognition is demonstrated in Figure \ref{fig:TL}.  

%% file: sections/results_WER_table_knowledge_transfer.tex
\begin{table*}[pos=h]
\centering
\caption{WER(\%) on ArzEn dev and test sets.}
\label{table:WER_knowledge_transfer}
\begin{adjustbox}{max width=\textwidth}
\begin{tabular}{|l|l|r|r|r|r|r|r|r|r|r|r|}
\hline
\multicolumn{1}{|c|}{\multirow{3}{*}{\textbf{\#Exp}}}&\multicolumn{1}{c|}{\multirow{3}{*}{\textbf{Datasets}}}&\multicolumn{6}{c|}{\textbf{Overall WER}}&\multicolumn{2}{c|}{\textbf{Arabic WER}}&\multicolumn{2}{c|}{\textbf{English WER}}
\\\cline{3-12}
\multicolumn{1}{|c|}{} & \multicolumn{1}{c|}{} & \multicolumn{2}{c|}{\textbf{GMM}}& \multicolumn{2}{c|}{\textbf{CNN-TDNN}}& \multicolumn{2}{c|}{\textbf{E2E}}& \textbf{CNN-TDNN}&\textbf{E2E}& \textbf{CNN-TDNN}&\textbf{E2E}\\\cline{3-12} 
\multicolumn{1}{|c|}{}& \multicolumn{1}{c|}{}& \multicolumn{1}{c|}{\textbf{dev}} & \multicolumn{1}{c|}{\textbf{test}} & \multicolumn{1}{c|}{\textbf{dev}} & \multicolumn{1}{c|}{\textbf{test}} & \multicolumn{1}{c|}{\textbf{dev}} & \multicolumn{1}{c|}{\textbf{test}}& \multicolumn{1}{c|}{\textbf{test}}& \multicolumn{1}{c|}{\textbf{test}}& \multicolumn{1}{c|}{\textbf{test}}& \multicolumn{1}{c|}{\textbf{test}} \\\hline
\multicolumn{12}{|c|}{\textbf{ [5.6h] ArzEn Data}}    \\\hline
1 & CS & \textbf{53.7} & \textbf{54.3} & 43.4 & 44.0 & 53.5 & 55.0 & 44.1 & 56.0 & 52.9 & 80.1 \\\hline
\multicolumn{12}{|c|}{\textbf{+ [28.4h] Egyptian Arabic Data (Callhome + MGB-3)}}    \\\hline
2 & CS+Eg & 54.4 & 55.1 & 35.1 & 35.8 & 33.6 & 35.3 & 35.6 & 32.9 & 44.6 & 63.5 \\\hline
\multicolumn{12}{|c|}{\textbf{+ [5-10h] English Data (Librispeech)}}   \\\hline
3 & CS+Eg+En(5h)  & 56.0 & 55.9  & 33.6 & 34.1  & 32.9 & 33.8 & 34.7 & 32.3 & 37.5 & 54.7 \\\hline
4 & CS+Eg+En(10h) & 57.6 & 57.4 & 32.9 & 33.2 & 32.1 & 33.0 & 33.8 & 31.6 & 36.3 & 51.8 \\\hline
\multicolumn{12}{|c|}{\textbf{+ [5-40h] MSA Data (MGB-2)}}       \\\hline
5 & CS+Eg+En(5h)+MSA(5h)  & 56.5 & 56.0 & 32.9 & 33.2 & \textbf{31.9}  & \textbf{32.8}  & 33.7 & 31.1 & 36.3 & 53.5  \\\hline
6 & CS+Eg+En(5h)+MSA(10h)  & 58.0 & 56.8 & 32.5 & 33.2 & 32.3 & 33.2 & 33.5 & 31.3 & 37.0 & 55.5 \\\hline
7 & CS+Eg+En(5h)+MSA(20h)  & 58.0 & 57.6 & 32.6 & 32.9 & 32.4 & 33.3 & 33.4 & 31.5 & 36.4 & 54.9  \\\hline
8 & CS+Eg+En(10h)+MSA(5h)  & 58.0 & 57.6 & 32.2 & 32.8 & 34.0 & 35.3 & 33.3 & 33.5 & 35.6 & 55.9  \\\hline
9 & CS+Eg+En(10h)+MSA(10h)  & 58.6 & 58.2 & 32.2 & 32.7 & 32.2 & 33.5 & 33.1 & 32.0 & 36.9 & 54.6 \\\hline
10 & CS+Eg+En(10h)+MSA(20h)  & 59.5 & 58.6 & \textbf{31.9} & \textbf{32.1}  & 32.8 & 34.2 & 32.6 & 32.3 & 34.2 & 51.6   \\\hline
11 & CS+Eg+En(10h)+MSA(40h)  & 59.5 & 59.0 & 31.7 & 32.6 & 35.8 & 37.1 & 33.0 & 35.9 & 36.4 & 58.3 \\\hline
\end{tabular}
\end{adjustbox}
\end{table*}

%% file: sections/results_WER_table_normalization.tex
\begin{table*}[pos=h]
\centering
\caption{Tolerance to Un-standardized Orthography: WER(\%) on ArzEn dev and test sets.}
\label{table:WER_normalization}
\begin{adjustbox}{max width=\textwidth}
\begin{tabular}{|l|p{5cm}|r|r|r|r|r|r|r|r|r|r|}
\hline
\multicolumn{1}{|c|}{\multirow{3}{*}{\textbf{\#Exp}}}&\multicolumn{1}{c|}{\multirow{3}{*}{\textbf{Datasets}}}&\multicolumn{6}{c|}{\textbf{Overall WER}}&\multicolumn{2}{c|}{\textbf{Arabic WER}}&\multicolumn{2}{c|}{\textbf{English WER}}
\\\cline{3-12}
\multicolumn{1}{|c|}{} & \multicolumn{1}{c|}{} & \multicolumn{2}{c|}{\textbf{GMM}}& \multicolumn{2}{c|}{\textbf{CNN-TDNN}}& \multicolumn{2}{c|}{\textbf{E2E}}& \textbf{CNN-TDNN}&\textbf{E2E}& \textbf{CNN-TDNN}&\textbf{E2E}\\\cline{3-12} 
\multicolumn{1}{|c|}{}& \multicolumn{1}{c|}{}& \multicolumn{1}{c|}{\textbf{dev}} & \multicolumn{1}{c|}{\textbf{test}} & \multicolumn{1}{c|}{\textbf{dev}} & \multicolumn{1}{c|}{\textbf{test}} & \multicolumn{1}{c|}{\textbf{dev}} & \multicolumn{1}{c|}{\textbf{test}}& \multicolumn{1}{c|}{\textbf{test}}& \multicolumn{1}{c|}{\textbf{test}}& \multicolumn{1}{c|}{\textbf{test}}& \multicolumn{1}{c|}{\textbf{test}} \\\hline
2 & CS+Eg & 54.4 & 55.1 & 35.1 & 35.8 & 33.6 & 35.3 & 35.6 & 32.9 & 44.6 & 63.5 \\\hline
12 & CS+Eg (Unstandardized) & 59.6 & 59.3 & 47.3 & 47.8 & 34.7 & 36.2 & 49.8 & 33.5 & 45.8 & 63.5 \\\hline
\end{tabular}
\end{adjustbox}
\end{table*}

%% file: sections/results_WER_table_CS_data.tex
\begin{table*}[pos=h]
\centering
\caption{Reliance on in-domain CS speech data: WER(\%) on ArzEn dev and test sets.}
\label{table:WER}
\begin{adjustbox}{max width=\textwidth}
\begin{tabular}{|l|p{5cm}|r|r|r|r|r|r|r|r|r|r|}
\hline
\multicolumn{1}{|c|}{\multirow{3}{*}{\textbf{\#Exp}}}&\multicolumn{1}{c|}{\multirow{3}{*}{\textbf{Datasets}}}&\multicolumn{6}{c|}{\textbf{Overall WER}}&\multicolumn{2}{c|}{\textbf{Arabic WER}}&\multicolumn{2}{c|}{\textbf{English WER}}
\\\cline{3-12}
\multicolumn{1}{|c|}{} & \multicolumn{1}{c|}{} & \multicolumn{2}{c|}{\textbf{GMMM}}& \multicolumn{2}{c|}{\textbf{CNN-TDNN}}& \multicolumn{2}{c|}{\textbf{E2E}}& \textbf{CNN-TDNN}&\textbf{E2E}& \textbf{CNN-TDNN}&\textbf{E2E}\\\cline{3-12} 
\multicolumn{1}{|c|}{}& \multicolumn{1}{c|}{}& \multicolumn{1}{c|}{\textbf{dev}} & \multicolumn{1}{c|}{\textbf{test}} & \multicolumn{1}{c|}{\textbf{dev}} & \multicolumn{1}{c|}{\textbf{test}} & \multicolumn{1}{c|}{\textbf{dev}} & \multicolumn{1}{c|}{\textbf{test}}& \multicolumn{1}{c|}{\textbf{test}}& \multicolumn{1}{c|}{\textbf{test}}& \multicolumn{1}{c|}{\textbf{test}}& \multicolumn{1}{c|}{\textbf{test}} \\\hline
\multicolumn{12}{|c|}{\textbf{Multi-condition Training}}       \\\hline
5 & CS+Eg+En(5h)+MSA(5h)  & 56.5 & 56.0 & 32.9 & 33.2 & \textbf{31.9}  & \textbf{32.8}  & 33.7 & 31.1 & 36.3 & 53.5  \\\hline
13 & Eg+En(5h)+MSA(5h)  & 65.3 & 64.2 & 54.2 & 53.0 & 54.5 & 54.7 & 51.1 & 54.4 & 101.0 & 113.8 \\\hline
\multicolumn{12}{|c|}{\textbf{Transfer Learning}}       \\\hline
14 & [Eg+En(5h)+MSA(5h)]\textrightarrow[CS]  & \cellcolor{Grey} & \cellcolor{Grey} & \cellcolor{Grey} & \cellcolor{Grey} & 32.8 & 34.3 & \cellcolor{Grey} & 32.6 & \cellcolor{Grey} & 56.1 \\\hline
\multicolumn{12}{|c|}{\textbf{Multi-condition Training + Transfer Learning}}       \\\hline
15 & [CS+Eg+En(5h)+MSA(5h)]\textrightarrow[CS]   & \cellcolor{Grey} & \cellcolor{Grey} & \cellcolor{Grey} & \cellcolor{Grey} & \textbf{31.6} & \textbf{32.7} & \cellcolor{Grey} & 31.1 & \cellcolor{Grey} & 53.7 \\\hline
\end{tabular}
\end{adjustbox}
\end{table*}

%% file: sections/qualitative_analysis.tex
\section{Qualitative Analysis}
\label{sec:qualitative}

We provide further insight into the systems' performances, identifying the strengths and weaknesses of both system and highlighting their complementary performance in the case of dialectal Arabic-English. We also discuss factors resulting in incorrect/harsh evaluation using WER.
\input{sections/examples_table_fine_tuned.tex}

\subsection{Hybrid Systems' Strength over End-to-end Systems}
\begin{figure}[h]
  \centering
  \includegraphics[width=\linewidth]{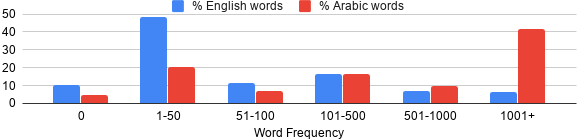}
  \caption{Percentage of words in test set against their word frequencies in train set}
  \label{fig:word_freq}
\end{figure}

The strength of TDNN-based systems is seen in their ability to recognize infrequent and OOV words that exist in the lexicon. This allows them to outperform E2E systems in extreme low-resource scenarios. This is shown in the significantly better recognition achieved by TDNN-based systems over E2E with models only trained on CS data (Table \ref{table:WER_knowledge_transfer}-Example 1). This is demonstrated in Table \ref{table:examples1}- Example 1 as well as in Table \ref{table:examples1}-Example 2, where \textit{deep} and \textit{learning} are infrequent words (in Librispeech and Arzen corpora) and \textit{reinforcement} is an OOV word. The superiority of TDNN-based systems in English recognition over E2E systems can be attributed to the low frequency of English words. As shown in Figure~\ref{fig:word_freq}, 49\% of English words in the test set occur only 1-50 times in train set (with 38\% occurring only 1-25 times), while 42\% of Arabic words occur 1000+ times. The limited amount of English data is found to negatively affect the performance of E2E systems' English recognition. DNN-based hybrid systems on the other hand are less affected by this problem as they compensate the lack of speech data by obtaining knowledge from lexicons. Accordingly, DNN-based systems show superiority in English recognition.
\vspace{1cm}

\subsection{End-to-end Systems' Strength over Hybrid Systems}
\input{sections/examples_WER_issues}
We identify two main scenarios where E2E systems' superiority is observed. Firstly, their recognition is not limited to lexicon words. This limitation disables TDNN-based systems to recognize frequently occurring words in the speech corpus that are absent in the lexicon. This strength is particularly advantageous in the case of dialectal Arabic as it allows E2E systems to better handle two main DA challenges: morphological richness and unstandardized orthography. Given that Arabic is a morphologically rich language, the limitation of lexicon dependence is highlighted, as it becomes more challenging to cover all words with all possible affixes combinations. This limits the performance of TDNN-based systems where (1) the system fails to recognize morphological compound words that are not present in the lexicon, even if they are frequently present in the speech corpus and (2) splitting of stems and prefixes in Arabic words is more frequently observed (for example ``\<الكتابة>'' meaning ``the writing'' was recognized as ``\<ال كتابة>''). This problem occurs when the lexicon includes each token separately, however the whole compound morphological word is absent. This limitation also affects English compound words, where they are recognized as stand-alone words (for example ``data base''). Although this issue does not occur frequently, it is to be noted as a disadvantage of the reliance on lexicons. The lexicon dependence of hybrid systems also limits their Arabic recognition in the case of orthographically unstandardized words. Spelling inconsistencies result in discrepancies between the speech corpora and the pronunciation lexicon, affecting lexicon OOV rates and recognition. The second strength point of E2E systems is their ability to recognize some OOV words with incorrect spelling, but very close pronunciation, such as \textit{coordinator}-\textit{courdinater} and \textit{spontaneously}-\textit{spontinuestly}. This is reflected in the lower Character Error Rate (CER) achieved on test set by the E2E system (18.4\%) in comparison to the TDNN-based system (22.0\%).

\input{sections/WER_issues}

%% file: sections/examples_table_fine_tuned.tex
\begin{table*}[pos=h]
  \caption{Hypotheses (test set) reflecting strengths and weaknesses of hybrid and E2E systems.}
  \label{table:examples1}
  \centering
  \setlength\tabcolsep{1.5pt}
  \begin{adjustbox}{max width=\textwidth}
  \begin{tabular}{|l|p{3.5cm}|r|}
    \hline
    \#&\multicolumn{1}{c|}{\textbf{Systems}}&\multicolumn{1}{c|}{\textbf{Hypothesis}}\\\hline
        &Ground-truth& 
        I'M NOT A TEAMS PERSON
        \<فلأ>
        \textleftarrow
        \\
        
        &Translation& 
        (so no I'M NOT A TEAMS PERSON)
        \\
        
        &CNN-TDNN[CS]&
        \textrightarrow
        NOT A TEAM PERSON
        \\
        
        &CNN-TDNN[+Eg]&
        \textrightarrow 
        I'M NOT A TEAM PERSON\\
        
        &CNN-TDNN[+Eg+En]&
        \textrightarrow 
        I'M NOT A TEAM PERSON\\
        
        1&CNN-TDNN[+Eg+En+MSA]&
        \textrightarrow 
        I'M NOT A TEAM PERSON\\
        
        &E2E[CS]&
        THINK PERSONAL
        \<لأ عندنا> \textleftarrow\\
        
        &E2E[+Eg]&
        I NOT A TEAM PERSON
        \<لأ> \textleftarrow \\

        &E2E[+Eg+En]&
        I NOT A TEAM PERSON
        \<لأ> \textleftarrow \\
        
        &E2E[+Eg+En+MSA]&
        I'M NOT A TEAM PERSON
        \<لأ> \textleftarrow\\
        
        \hline
        
        
        &Ground-truth& 
        \underline{DEEP REINFORCEMENT LEARNING}
        \<اسمه>
        TOPIC
        \<بتاعى عن>
        BACHELOR PROJECT 
        \<ال> \textleftarrow\\
        
        &Translation& 
        (my BACHELOR PROJECT is about a TOPIC called DEEP REINFORCEMENT LEARNING)\\
       
        &CNN-TDNN[+Eg+En]&
        LEARNING
        \<من ال>
        \underline{DIP DREAM FORCE}
        \<اسمه>
        TOPIC
        \<بتاعى عن>
        BACHELOR PROJECT
        \<و ال> \textleftarrow\\
        
        2&CNN-TDNN[+Eg+En+MSA]&
        \underline{DEEP REINFORCEMENT LEARNING}
        \<اسمه>
        TOPIC 
        \<بتاعى عن>
        BACHELOR PROJECT 
        \<ال> \textleftarrow\\

        &E2E[+Eg+En]&
        \underline{
        LEARNING
        \<ال> 
        DEEP REFORCEMENT}
        \<اسمه> 
        TOPIC
        \<بتاعى عن>
        BACHELOR PROJECT
        \<ال> \textleftarrow\\
        
        &E2E[+Eg+En+MSA]&
        \underline{PERIENFORCEMENT IN LEARNING
        \<دى>}
        \<اسمه>
        TOPIC
        \<بتاعى عن>
        BACHELOR PROJECT
        \<ال>
        \textleftarrow\\
        
        &Transfer Learning&
        \underline{PERIENFORCEMENT LEARNING
        \<دى>}
        \<اسمه>
        TOPIC
        \<بتاعى عن>
        BACHELOR PROJECT
        \<ال>
        \textleftarrow\\\hline
        
  \end{tabular}
  \end{adjustbox}
\end{table*}

%% file: sections/examples_WER_issues.tex
\begin{table*}[pos=h]
  \caption{Hypotheses (test set) reflecting issues with WER as an evaluation metric for the Egyptian Arabic-English language pair.}
  \label{table:examples2}
  \centering
  \setlength\tabcolsep{1.5pt}
  \begin{adjustbox}{max width=\textwidth}
  \begin{tabular}{|l|p{3.5cm}|r|}
    \hline
    \#&\multicolumn{1}{c|}{\textbf{Systems}}&\multicolumn{1}{c|}{\textbf{Hypothesis}}\\\hline        
    
        &Ground-truth& 
        \underline{\<موبايل>}\<ال>
        \<اهم م>
        LAPTOP
        \<انا بالنسبالى ال >
        \textleftarrow\\
        
        &Translation& 
        (for me the LAPTOP is more important than the MOBILE)\\
       
        1&CNN-TDNN[+Eg+En+MSA]&
        \underline{MOBILE}
        \<اهم من ال >
        LAPTOP
        \<انا بالنسبة لى ال>
        \textleftarrow\\ 
        
        &E2E[+Eg+En+MSA]&
        \underline{MOBILE}
        \<اهم من ال >
        LAPTOP
        \<انا بالنسبالى ال>
        \textleftarrow\\\hline   
                
        &Ground-truth& 
        BACHELOR
        \<عشان ال> 
        \underline{\<دلوقت>}
        ENGLISH
         \<كتير بال>
         PAPERS
        \<بقرا>
        \<كتاب قبل كده و>
        \underline{\<قاريت> } \textleftarrow\\
        
        &Translation&
        (I read a book before and I'm reading a lot of PAPERS now for the BACHELOR)\\
         
        2&CNN-TDNN[+Eg+En+MSA]&
        BACHELOR
        \<عشان ال> 
        \underline{\<دلوقتى>}
        ENGLISH
        \<كتير بال>
         PAPERS
        \<كتاب قبل كده و>
        \underline{\<قريت> } \textleftarrow\\
        
        &E2E[+Eg+En+MSA]&
        BACHELOR
        \<عشان> 
        \underline{\<دلوقت>}
        ENGLISH
        \<كتير بال>
         PAPER
        \<بقرا>
        \<كتاب قبل كده و>
        \underline{\<قريت> } \textleftarrow\\\hline
        &Ground-truth& 
        \<هناك>
        \underline{\<حاسافر>}
        \<فعلا لما يبقى عندى فرصة تانية انا لازم>
        ACTUALLY
        \<و> \textleftarrow\\

        &Translation&
        (and ACTUALLY truly when I get another opportunity I must go there)\\
        
        3&CNN-TDNN[+Eg+En+MSA]&
        \underline{\<اسافر>}
        \<فعلا لما يبقى عندى فرصة تانية انا لازم>
        ACTUALLY
        \<و> \textleftarrow\\
        
        &E2E[+Eg+En+MSA]&
        \<هناك>
        \underline{\<اسافر>}
        \<فعلا لما يبقى عندى فرصة تانية انا لازم>
        ACTUALLY
        \<و> \textleftarrow\\\hline
  \end{tabular}
  \end{adjustbox}
\end{table*}

%% file: sections/WER_issues.tex
\subsection{Incorrect/Harsh Evaluation Using WER}
We identify cases where WER acts as an incorrect or harsh evaluation metric for dialectal Arabic-English ASR recognition. Firstly, the confusion between loanwords and single-word switches. 
Loanwords are single words that are borrowed from one language and become, possibly adapted, and integrated into another \cite{Has09}, such as \textit{doctor}-\<دكتور> (d-u-k-t-o-o-r) and \textit{Paris}-\<باريس> (p-a-r-e-e-s/b-a-r-e-e-s). It is a difficult task for humans to distinguish between both multilingual phenomena, especially with CS words often pronounced with foreign accent, resulting in three cases: loan words and accented/non-accented single-word switches. It is interesting to see how our ASR systems handle the distinction between loanwords and single-word switches and their ability to use the correct orthography in each case; Arabic for loanwords, and Latin for CS. Our models are mostly able to transcribe loanwords and non-accented single-word switches with the correct orthography. The confusion however arises with accented single-word switches. We also note that in some cases, loanwords have the same pronunciation in both languages, such as \textit{mobile}-\<موبايل>. 
In this case, both transcriptions are acceptable, and the confusion between both incorrectly affects WER, as shown in Table \ref{table:examples2}-Example 1.
Secondly, due to the issue of unstandardized orthography, correctly recognized words having a different orthography than the reference words are incorrectly penalized. This is seen in Table \ref{table:examples2}-Example 2, where underlined words are all acceptable spellings.
Thirdly, with Arabic being a morphologically rich language, words can be combined with several affixes, and consequently, each stem produces numerous words with minor differences, which poses a challenge for ASR systems. WER then acts as a harsh evaluation metric in the case of morphological compound words, where words with a mistake in one affix are considered to be incorrect. As shown in Table \ref{table:examples2}-Example 3, the word ``\<حاسافر>'' (will travel) was recognized as ``\<اسافر>'' (to travel). In this case, the future proclitic ``\<ح>'' that was dropped in the hypothesis does not affect the meaning of the sentence as the tense is already clear from the sentence context, and the use of both words in this example is correct. These factors contribute to a significantly lower CER, where our best models achieve CER of 22.0\%/18.4\% (CNN-TDNN/E2E) on test set, in comparison to WER of 32.1\%/32.7\% (CNN-TDNN/E2E).

%% file: sections/merging_hyp.tex
\section{Merging Hypotheses}
Motivated by the different strengths provided by each system, we propose and evaluate combination approaches for hypotheses on both levels: sentence and word. We report the results of the combination approaches in Table \ref{table:merge_wer}.

\begin{table}[pos=h]
\centering
\caption{WER improvements achieved on dev and test sets.}
\label{table:merge_wer}
\setlength\tabcolsep{1.5pt}
\begin{adjustbox}{max width=\textwidth}
\scalebox{0.95}{
\begin{tabular}{|l|r|r|r|r|}
\hline
\multicolumn{1}{|c|}{\multirow{2}{*}{\textbf{Model}}} & \multicolumn{2}{c|}{\textbf{Overall WER}} & \textbf{Arabic WER} & \textbf{English WER} \\ \cline{2-5} 
 & \multicolumn{1}{c}{\textbf{dev}} & \multicolumn{1}{|c}{\textbf{test}} & \multicolumn{1}{|c}{\textbf{test}} & \multicolumn{1}{|c|}{\textbf{test}} \\\hline
CNN-TDNN & 31.9 & 32.1 & 32.6 & 34.2 \\ \hline
E2E & 31.6 & 32.7 & 31.1 & 53.7 \\ \hline
\multicolumn{5}{|c|}{\textbf{Sentence-level Combination}}\\\hline
Oracle WER & 26.5  & 26.9 & 26.5 & 35.8 \\ \hline
Confidence Scores & 31.1 & 32.0 & 31.4 & 43.3 \\ \hline
Discriminant Classifier & 30.8 & 31.0 & 30.9 & 38.5 \\ \hline
\multicolumn{5}{|c|}{\textbf{Word-level Combination}}\\\hline
E2E \textrightarrow CNN-TDNN & 31.4 & 31.5 & 31.9 & 34.1 \\ \hline
CNN-TDNN \textrightarrow E2E & 31.0 & 32.0 & 31.1 & 48.8 \\ \hline
\multicolumn{5}{|c|}{\textbf{Hybrid Combination}}\\\hline
sentence+word level & \textbf{30.4} & \textbf{30.6}& \textbf{30.6} & \textbf{37.2} \\ \hline
\end{tabular}
}
\end{adjustbox}
\end{table}

\subsection{Sentence-level Combination}
In this approach, we propose merging the results of both systems through rescoring. In order to validate the possible improvements that can be achieved, we check the oracle WER. The Oracle WER is obtained by choosing the hypothesis with lower WER among the 1-best hypotheses from both systems. The oracle WER presents an indicator for the upper bound improvement that can be achieved through sentence-level systems combination. As shown in Table \ref{table:merge_wer}, there is 5.2-5.8\% absolute WER difference between the TDNN-based and E2E systems and Oracle WER. This supports our claim that both systems contribute in providing the best hypothesis.

\subsubsection{Confidence Scores}
In the first combination method, we rely on the confidence scores assigned to the 1-best hypotheses from both systems, to choose the hypothesis with the highest score. ESPNET provides one confidence score per hypothesis, while Kaldi provides separate scores from acoustic and language models, where scores are unnormalized log probabilities. For each utterance, we obtain comparable normalized probabilities for 1-best hypotheses by applying Softmax function to the N-best (N=20) hypotheses. The final confidence score of hypothesis $i$ is calculated as follows:
  \begin{align}
    &score_{ESPNET} = \max_{1 \leq i \leq N}Softmax(\textbf{e}_i)\\
    &score_{KALDI} = \max_{1 \leq i \leq N}Softmax(\textbf{k}_i)\\
    &\textbf{k}_i=(-1\times\textbf{lm}_i)+(-1\times\textbf{am}_i\times\frac{1}{w}) \\
    &Softmax(\textbf{z}_i)= \frac{exp(\textbf{z}_i)}{\sum_{j=1}^{N}exp(\textbf{z}_j)} 
  \end{align}
where $\textbf{e}$ and $\textbf{k}$ are ESPNET and Kaldi scores' vectors of 20-best hypotheses for the utterance, respectively. For Kaldi, the overall confidence score of hypothesis $i$ in the N-best hypotheses is calculated from the separate language model $\textbf{lm}_i$ and acoustic model $\textbf{am}_i$ scores. In this calculation, the acoustic score needs to first be scaled down by $w$, which is the LM weight for the best WER ($w=8$ for our best Kaldi model). After obtaining sentence confidence scores for the best hypothesis from each system, we choose the hypothesis with the higher score, giving slight absolute WER improvements of 0.1(CNN-TDNN)-0.7(E2E)\% on test set, as shown in Table \ref{table:merge_wer}.

\subsubsection{Discriminant Classifier}
The confidence score-based approach allows the decision to rely on one factor only; the confidence score. However, other factors can also play an important role, such as sentence CMI and the percentage of English words in the sentence. Therefore, we propose treating the problem as a Binary Classification problem. Given the following information about 1-best hypotheses; confidence scores, CMI and English percentage, we train a Discriminant Classifier to learn which system to choose. The model achieves an accuracy of 62.1\% and improves WER by 1.1(CNN-TDNN)-1.7(E2E)\% (absolute) and 3.4(CNN-TDNN)-5.2(E2E)\% (relative) on test set, as shown in Table \ref{table:merge_wer}. We also experiment with $n>1$, however, we observe a reduction in the accuracy of the Discriminant Classifier for $n>1$ and an increase in WER. In Table \ref{table:ML_n_results2}, we show the results of training the Discriminant Classifier for n=1,2,5. We report the accuracy of the Discriminant Classifier as well as the percentage of hypotheses chosen from TDNN-based and E2E systems and the corresponding WER on test set. 

\begin{table}[pos=h]
\centering
\caption{Results of sentence-level combination using Discriminant Classifier for n=1,2,5.}
\label{table:ML_n_results2}
\begin{tabular}{|l|r|r|r|r|r|r|r|}
\hline
\textbf{n} & \multicolumn{1}{c|}{\textbf{\textbf{ML Accuracy}}} & \multicolumn{1}{c|}{\textbf{\textbf{\% CNN-TDNN}}} & \multicolumn{1}{c|}{\textbf{\textbf{\% E2E}}} & \multicolumn{1}{c|}{\textbf{\textbf{WER}}} \\ \hline
n=1 & 62.1 & 41.6 & 58.4 & 31.0 \\\hline
n=2 & 34.0 & 26.5 & 73.5 & 31.8 \\ \hline
n=5 & 31.6 & 33.0 & 67.0 & 32.5\\\hline
\end{tabular}
\end{table}

\begin{table*}[h]
\caption{Effect of merging approaches (test set example).}
\label{table:merge_example}
\setlength\tabcolsep{1.5pt}
\begin{adjustbox}{max width=\textwidth}
\begin{tabular}{|l|r|r|r|}
\hline
 \multicolumn{1}{|c|}{\textbf{Model}}& \multicolumn{1}{c|}{\textbf{Sentence}} & \multicolumn{1}{c|}{\textbf{Score}} & \multicolumn{1}{c|}{\textbf{WER}} \\\hline
reference & 
\<معين فى دماغى دلوقتى>
QUOTE
\<يعنى بس انا مافيش>
STANDARDS
\<هو فى>
& \cellcolor{Grey} & \cellcolor{Grey} \\\hline
CNN-TDNN &
\textcolor{red}{\underline{\<يعنى>}}
\textcolor{blue}{\underline{\<دلوقت>}}
\<معين فى دماغى>
QUOTE
\textcolor{blue}{\underline{\<ما فيش>}}
\<يعنى بس انا >
STANDARDS
\<فى>
 \<هو>
& 0.20 & 0.33\\\hline
E2E & 
\<معين فى دماغى دلوقتى>
QUOTE
\<يعنى بس انا مافيش>
\textcolor{red}{\underline{STUDENTS}}
\<هو فى>
& 0.43 & 0.08\\\hline
\begin{tabular}[c]{@{}l@{}}sentence-level\\ (E2E chosen)\end{tabular} &
\<معين فى دماغى دلوقتى>
QUOTE
\<يعنى بس انا مافيش>
\textcolor{red}{\underline{STUDENTS}}
\<هو فى>
& 0.43 & 0.08\\\hline
+word-level &
\<معين فى دماغى دلوقتى>
QUOTE
\<يعنى بس انا مافيش>
STANDARDS
\<هو فى>
& \cellcolor{Grey} & 0.0\\\hline
\end{tabular}
\end{adjustbox}
\end{table*}

\begin{table*}[H]
\centering
\caption{WER improvements achieved on test set with regards to CS types.}
\label{table:merge_cs_wer}
\setlength\tabcolsep{1.5pt}
\begin{adjustbox}{max width=\textwidth}
\begin{tabular}{|p{3.5cm}|r|r|r|r|r|r|r|r|r|r|}
\hline
\multicolumn{1}{|c|}{\multirow{2}{*}{\textbf{Model}}} &
\multicolumn{3}{c|}{\textbf{All sentences}} &
\multicolumn{3}{c|}{\textbf{Intra-sentential CS}} &
\multicolumn{3}{c|}{\textbf{Intra-word CS}}\\\cline{2-10} 
&\textbf{overall} & \multicolumn{1}{c|}{\textbf{Eg}} & \multicolumn{1}{c|}{\textbf{En}} & \textbf{Overall} & \multicolumn{1}{c|}{\textbf{Eg}} & \multicolumn{1}{c|}{\textbf{En}} & \textbf{Overall} & \textbf{Eg affixes} & \textbf{En words} \\\hline
CNN-TDNN & 32.1 & 32.6 & \textbf{34.2} & 31.5 & 32.2 & \textbf{32.8} & \textbf{32.6} & 37.2 & \textbf{29.8}\\ \hline
E2E & 32.7 & 31.1 & 53.7 & 32.9 & 31.0 & 52.7 & 39.2 & \textbf{28.8} & 48.8\\\hline
Hybrid Combination & \textbf{30.6} & \textbf{30.6} & 37.2 & \textbf{30.0} & \textbf{30.0} & 35.8 & 35.2 & 38.1 & 34.7 \\\hline
\end{tabular}
\end{adjustbox}
\end{table*}
\subsection{Word-level Combination}
In this approach, we take advantage of the strengths of both systems to merge top hypotheses on the word level. We aim at utilizing the English recognition provided by TDNN-based systems, and overcome the limitation of TDNN-based systems' of failing at recognizing words that are not present in the lexicon. We investigate the effect of (1) borrowing English words from CNN-TDNN hypotheses into ESPNET hypotheses as well as infrequent and OOV Arabic words present in lexicon and (2) borrowing Arabic words that are not present in the lexicon from E2E hypotheses into CNN-TDNN hypotheses. We first obtain alignments of 1-best hypotheses using fast-align \cite{DWS+10}, and perform the replacements accordingly. While replacing words having 1-to-many alignments deteriorated the WER, replacing 1-to-1 alignments gave WER improvements, as shown in Table \ref{table:merge_wer}. By looking into the replacements, we find that this approach improves English E2E recognition by inserting OOV words in lexicon recognized by TDNN-based models (\textit{please} \textrightarrow \textit{peers} and \textit{interpanered} \textrightarrow \textit{entrepreneur}), as well as correcting mis-spelt words (\textit{alcommist} \textrightarrow \textit{alchemist} and \textit{chaveling} \textrightarrow \textit{traveling}). On the other hand, TDNN-based hypotheses were improved by inserting more accurate words provided by the E2E model where affixes were corrected (\<هواية> (hobby) \textrightarrow \<هوايات> (hobbies) and \<صاحبتى> (my friend) \textrightarrow \<صاحبتها> (her friend), \<دخلت> (entered) \textrightarrow \<دخلته> (entered it)). This further validates the strength of E2E systems over hybrid systems in the case of morphologically rich languages. The word-level combination resulted in a 1.9\% and 2.1\% relative improvements for TDNN-based and E2E systems.

\subsection{Hybrid Combination}
In the hybrid combination, the merged 1-best hypotheses list obtained from sentence-level combination is used as input to word-level combination. In Table \ref{table:merge_example}, we show an example for the effect of sentence-level and word-level merge approaches, where words colored in red are incorrect and words colored in blue are correct however differ from the reference due to unstandardized orthography. As shown in Table \ref{table:merge_wer}, the final model achieves absolute WER improvements of 2.0\% on Arabic over TDNN-based model and 16.5\% on English over E2E models. Relative overall improvements of 4.7\% and 6.4\% are achieved over TDNN-based and E2E systems. We have also performed hybrid combination for $n>1$, however, it does not outperform results for $n=1$, where WER on test set drops to 31.3\% (n=2) and 32.1\% (n=5).\\

We further evaluate the effect of our hybrid merging approach on the ASR system performance with regards to code-switching. We report the WER for intra-sentential and intra-word CS. We report the WER figures in Table \ref{table:merge_cs_wer}. In the case of intra-sentential CS, we report WER for code-mixed sentences only. We show that the hybrid combination achieves a relative improvement of 4.8\% and 8.8\% over TDNN-based hybrid and E2E systems for code-mixed sentences. In order to calculate WER for morphological CS words, we use fast-align \cite{DCS13} to obtain word level alignments between the tokens making up the compound morphological CS word in the reference transcription and their corresponding tokens in the merged hypothesis. The tokens were then manually revised to fix alignment mistakes. We report that in the case of intra-word CS, E2E system is superior in recognizing Arabic prefixes, while TDNN-based system is superior in recognizing English embedded words. For overall recognition of morphological code-switched words, TDNN-based system shows superiority over E2E and combined systems. Through our error analysis, we observe that similar to words, TDNN-based systems were unable to recognize the following Arabic affixes \{
\<ي>,
\<هت>,
\<كال>,
\<بى>,
\<م>,
\<فب>,
\<هن>,
\<ات>,
\<وا>,
\<ى>,
\<ها>,
\<نى>\} as most of these affixes are not present in the lexicon, while E2E systems were not able to recognize the following infrequent affixes: \{
\<هت>,
\<كال>,
\<م>,
\<فب>,
\<هن>,
\<ات>,
\<وا>,
\<ها>,
\<نى>
\}.  


%% file: sections/conclusion_v2.tex
.\section{Conclusions}
In this paper, we develop the first Egyptian Arabic-English CS ASR system. This is a challenging task due to the scarcity of CS corpora, the fact that the Egyptian Arabic primary language is under-resourced as well as Dialectal Arabic issues. Our work aims at filling the gap in resources and NLP research for Dialectal Arabic-English code-switching. We present the ArzEn speech corpus, provide CS analyses, and release the corpus to motivate further NLP and linguistic research for this language pair. We extend CS ASR research by presenting a thorough comparison between TDNN-based hybrid and Transformer-based end-to-end systems under the setting of this low-resourced, morphologically rich, orthographically unstandardized, code-switched language pair. We compare both systems with regards to their extent of multilingual and crosslingual knowledge transfer, their reliance on in-domain CS data, and their tolerance to unstandardized orthography. In general, E2E systems demonstrate great potential, providing comparable results to TDNN-based systems. We also show that end-to-end and hybrid systems both offer complementary sets of strength points. On the one hand, E2E systems are robust against Dialectal Arabic issues, outperforming TDNN-based systems in Arabic recognition. On the other hand, TDNN-based systems show significant superiority in English recognition. We therefore propose system combination as an effective method for leveraging the strengths of both systems through word- and sentence-level hypotheses combination. Our results prove the complementarity of both systems, where system combination achieves WER relative improvements of 4.7\% in overall recognition and 4.8\% in the case of code-mixed sentences. By looking into the Oracle WER for sentence-level combination, we show that further gains can be achieved, and thus we motivate further research in this direction. Our final ASR system achieves 30.6\% WER and 18.7\% CER on ArzEn test set.

%% file: sections/intrawordCS_table.tex
\section{Arzen Intra-word CS}
\label{sec:appendix-morphCS}
In Table \ref{table:arzen_affixes}, we provide a list of the affixes and clitics occurring in morphological CS words in ArzEn corpus, along with their frequencies.

\begin{table*}[H]
\centering
\begin{adjustbox}{max width=0.9\textwidth}
\begin{tabular}{|l|l|l|r|l|}\hline
\textbf{Affix} & \textbf{Meaning} & \textbf{Frequency} & \textbf{Examples} & \textbf{Translation} \\\hline
\multicolumn{5}{|c|}{\textbf{Proclitics}} \\\hline
\<ال> &  Definite article & 1,995 &WEEKENDS+\<ال> & the weekends \\\hline
\<ف> & 
Conjunction `so/then' & 83 & THEREFORE+\<ف> & so therefore \\\hline
 \<ك> & Preposition `as' & 60 & STUDENT+\<ك> & as (a) student \\\hline
 \<ل> & Preposition `to/for' & 45 & LEVEL+\<ل> & to (a) level \\\hline
 \<ب> & Preposition `with' & 36 &ATTITUDE+\<ب>& with (an) attitude \\\hline
 \<ه> or \<ح> & 
Imperfect future verb & 7 & SUBMIT+\<ه> & will submit \\\hline
\multicolumn{5}{|c|}{\textbf{Enclitics}} \\\hline
\<ه> & 
3rd person possessive pronoun (masc.) & 4 & 
\multicolumn{1}{|l|}{CAREER\#\<ه>}
&his career  \\\hline
\<ها> &
3rd person direct object pronoun (fem.) & 1 & \<ها>\#PROPAGATE+\<ي> & (he) propagates it\\\hline
\<ى> & 
1st person possessive pronoun  & 1 & 
\multicolumn{1}{|l|}{MOBILE\#\<ى>}
& my mobile \\\hline
\multicolumn{5}{|c|}{\textbf{Prefixes}} \\\hline
 \<ا>& 
 1st person imperfect verb (sg.) & 71 & IMPLEMENT+\<ا> & to implement \\\hline
\<ي> & 
3rd person imperfect verb (masc.) & 49 & BREAK+\<ي> & to break \\\hline
 \<ب> & Present tense marker & 46 & USE+\<ب> & to use \\\hline
\<ت> & 
2nd person imperfect verb (masc.) & 33 & PARTICIPATE+\<ت> & to participate \\\cline{2-2}\cline{4-5}
 & 
3rd person imperfect verb (fem.) &  & SUPPORT+\<ت> & to participate \\\hline
\<ن> & 
1st person imperfect verb (pl.) & 14 & DETECT+\<ن> & to detect \\\hline
\<م>& Present participle & 1 & EXPECT+\<م> & EXPECTING \\\hline
\multicolumn{5}{|c|}{\textbf{Suffixes}} \\\hline
\<ات> & Feminine plural & 7 & 
\multicolumn{1}{|l|}{EVENT\#\<ات>}
& events \\\hline
\<نى> & 
1st person direct object pronoun  & 1 & \<نى>\#EVALUATE+\<بي> & he evaluates me\\\hline
\multicolumn{5}{|c|}{\textbf{Circumfixes}} \\\hline
\<وا> + \<ي>& 
3rd person imperfect verb (pl.)  & 4 & \<وا>\#ADAPT+\<ي> & (they) adapt\\\hline
\<ى> + \<ت>& 
2rd person imperfect verb (fem.)  & 2 & \<ى>\#CODE+\<ت> & to code\\\hline
\<وا> + \<ت>&
2rd person imperfect verb (pl.)  & 1 & \<وا>\#PRESENT+\<ت> & to present\\\hline
\multicolumn{5}{|c|}{\textbf{Combinations}} \\\hline
\<بال> & Combination of the morphemes: \<ب> (preposition) and \<ال>
& 85 & Mobile+\<بال> & with the mobile \\\hline
\<لل> & Combination of the morphemes: \<ل> and \<ال> & 36 &LIMIT+\<لل> & to the limit  \\\hline
\<بي> & 
Combination of the morphemes: \<ب> (present tense) and \<ي> & 27 & APPRECIATE+\<بي> & (he/it) appreciates \\\hline
\<بت> & 
Combination of the morphemes: \<ب> (present tense) 
& 19 & OVERCOME+\<بت> & (you) overcome \\
& and \<ت> (2nd person imperfect verb (masc.))&  & & \\\cline{2-2}\cline{4-5}
 & 
Combination of the morphemes: \<ب> (present tense)
&  & REQUIRE+\<بت> & (she/it) requires \\
 & 
and \<ت> (3rd person imperfect verb (fem.))
&  & &  \\\hline
\<بن> & 
Combination of the morphemes: \<ب> (present tense) and \<ن>
& 10 & TARGET+\<بن> & (we) target \\\hline
\<فال> & Combination of the morphemes: \<ف> and \<ال> & 6 & TEAM+\<فال> & so the team \\\hline
\<هي>& 
Combination of the morphemes: \<ه> and \<ي> & 6 & REACT+\<هي> & 
(he/it) will react \\\hline
\<هت> & 
Combination of the morphemes: \<ه> and \<ت>
& 6 & AFFECT+\<هت> & 
(she/it) will affect \\\hline
\<كال> 
& Combination of the morphemes: \<ك> and \<ال> & 2 & WORKING LIFE+\<كال> & as (the) working life \\\hline
\<فب>& 
Combination of the morphemes: \<ف> and \<ب> (present tense)
& 2 & GROW+\<فب> & so (I) grow  \\\hline
\<هن>& 
Combination of the morphemes: \<ه> and \<ن>
& 2 & GUARANTEE+\<هن> & (we) will guarantee \\\hline
\<وا> + \<بي>& 
Combination of the morphemes: \<ب> (present tense) and \<وا> + \<ي> circumfix
& 2 & \<وا>\#target+\<بي> & (they) target\\\hline
\<ى> + \<بت>& 
Combination of the morphemes: \<ب> (present tense) and \<ى> + \<ت> circumfix
& 1 & \<ى>\#HANDLE+\<بت> & to handle\\\hline
\end{tabular}
\end{adjustbox}
\caption{Affixes and clitics occurring in ArzEn Intra-word CS.}
\label{table:arzen_affixes}
\end{table*}